\documentclass{article}
\usepackage{amsmath,amsthm,amssymb,graphicx,mlspconf}
\usepackage[ruled,vlined,linesnumbered,lined,boxed]{algorithm2e}
\RequirePackage[colorlinks,citecolor=blue,urlcolor=blue,breaklinks=true]{hyperref}

\usepackage{tikz}
\usepackage{pgfplots}
\pgfplotsset{compat=1.5}
\let\oldnl\nl% Store \nl in \oldnl
\newcommand{\nonl}{\renewcommand{\nl}{\let\nl\oldnl}}% Remove line number for one line
\copyrightnotice{U.S.\ Government work not protected by U.S.\ copyright}

\copyrightnotice{978-1-7281-6662-9/20/\$31.00 {\copyright}2020 Crown}

\copyrightnotice{978-1-7281-6662-9/20/\$31.00 {\copyright}2020 European Union}

\copyrightnotice{978-1-7281-6662-9/20/\$31.00 {\copyright}2020 IEEE}

\toappear{2020 IEEE International Workshop on Machine Learning for Signal Processing, Sept.\ 21--24, 2020, Espoo, Finland}

\title{Block-wise Minimization-Majorization algorithm for Huber's criterion:
sparse learning and applications} 
\name{Esa Ollila and Ammar Mian\thanks{This work was supported by the Academy of Finland
under Grant 298118.}}
\address{Department of Signal Processing and Acoustics, Aalto University, Finland}
\newcommand{\ndim}{N}
\newcommand{\bo}[1]{\mathbf{#1}}

\newcommand{\X}{\bo X}

\newcommand{\x}{\bo x}
\newcommand{\y}{\bo y}
\renewcommand{\r}{\bo r}  % residual vector   
\newcommand{\xr}[1]{x_{#1}}%{\x_{#1 {\displaystyle \cdot} }}    % row vector of MM 

\DeclareMathOperator*{\argmin}{arg\,min} % and min

\newcommand{\R}{\mathbb R} % field of real numbers
\newcommand{\E}{\expec}
\newcommand{\iidsim}{\overset{i.i.d.}{\sim}}

\newcommand{\expec}{\mathsf{E}}
\newcommand{\pr}{\partial}
\newcommand{\sig}{\sigma}
\newcommand{\al}{\alpha} 
\newcommand{\lam}{\lambda} 
\newcommand{\be}{\beta} % \bom{\beta}}

\newcommand{\bmat}{\begin{pmatrix}}
\newcommand{\emat}{\end{pmatrix}}
\newcommand{\beq}{\begin{equation}}
\newcommand{\eeq}{\end{equation}}

\newcounter{ctheorem}
\newtheorem{theorem}[ctheorem]{Theorem}

\begin{document}
%\ninept

\maketitle

\begin{abstract}
Huber's criterion can be used for robust joint estimation of regression and scale parameters in the linear model. 
Huber's \cite{huber1981robust}  motivation for introducing the criterion stemmed from non-convexity of the joint maximum likelihood objective function  as well as non-robustness (unbounded influence function) of the associated ML-estimate of scale. 
In this paper, we illustrate how the original algorithm proposed by Huber can be set within the block-wise minimization majorization framework. In addition, we propose novel data-adaptive step sizes for both the location and scale, which are further improving the convergence. We then illustrate how Huber's criterion can be used for sparse learning of underdetermined linear model 
using the iterative hard thresholding approach. We illustrate the usefulness of the algorithms in an image denoising application and simulation studies. 
\end{abstract}
%\cite
%

%-- Section 1 ---
\section{Introduction} \label{sec:intro}

Consider having  $\ndim$ measurements or {\it outputs} (responses)  $y_i \in \R$   
and each output  is associated with a  $p$-dimensional vector  of {\it inputs} (predictors) 
$\xr{i}^\top = (x_{i1}, \ldots, x_{ip}) \in \R^p$.  
We assume a linear regression model, where input-output relationship is described by  
\beq \label{eq:linmodel} 
y_i = \xr{i}^\top \be + e_i   , \quad i=1,\ldots,\ndim,  
\eeq 
where the  random error terms $e_i$, $i=1,\ldots,\ndim$,  are independent and identically distributed (i.i.d.) and account for both the modelling and measurement errors. 
The distribution of the errors is assumed to be symmetric with probability density function (p.d.f.) $f(e)=(1/\sigma) f_0(e/\sigma)$, where $\sigma$ denotes the scale parameter of the distribution 
and $f_0(e)$ denote the standardized (unit scale) distribution. 
The goal is to  estimate the vector  $\be =(\beta_1,\ldots,\beta_p)^\top  \in \R^p$ 
 of  \emph{regression coefficients}  and the \emph{scale} parameter $\sigma$ given the data %$ \mathcal Z = \{   \z_i= 
 $\{ (y_i,\xr{i}) ; i=1, \ldots,\ndim\}$.  
 In many applications,  the scale is  a nuisance parameter, and the primary interest is on estimation of $\be$.

The linear model can be conveniently expressed using matrix-vector notations.  
We use the convention that matrices are represented by bold uppercase letters while 
vectors will not be bold, except when they have $N$ components. 
We may then express the set of $N$ input $p$-vectors compactly via the $N \times p$ design matrix,  
$\X  = \bmat \xr{1} &  \xr{2} & \ldots &  \xr{\ndim} \emat^\top 
=\bmat \ \,  \bo x_1  & \  \ \bo x_2 \ & \  \cdots &  \bo x_{p} \ \, \emat $. 
The convention distinguishes a $p$-vector of inputs $x_i$ for the $i^{th}$ observation from the $N$-vectors $\bo x_i$ consisting of all the observations on $i^{th}$ variable. 
We collect the outputs into a vector $\bo y=(y_1,\ldots,y_N)^\top$ and error terms into vector $\bo e$. Thus the linear model \eqref{eq:linmodel} now rewrites as 
$
\y = \X \be + \bo e$.

We return to the problem of joint robust estimation of regression and scale parameters in the linear model using 
Huber's  criterion \cite{huber1981robust}.
 In this paper, we elaborate on  recent derivations in \cite{zoubir2018robust} showing how the original algorithm proposed by Huber can be derived as a block-wise minimization majorization algorithm \cite{razaviyayn2013unified}.    In addition, we propose new novel data-adaptive step sizes for both the location and scale parameter updates, which are illustrated to  improve the convergence.  We then describe the use of Huber's criterion in sparse  signal recovery (SSR) \cite{donoho2006stable,duarte2011structured} problem using normalized iterative hard thresholding (NIHT) approach  \cite{ollila2014robust,ollila2015robust}. Robust performance of  Huber's sparse and non-sparse estimates of regression and scale are illustrated with a simulation study and in robust image denoising application and Matlab and Python toolboxes are made available in \verb+github.com/AmmarMian/huber_mm_framework+. We hope that the present  paper  and the toolboxes are able to bring more attention to Huber's criterion which is scalable robust approach for many practical large-scale problems. Finally, we note that sparse/non-sparse linear estimation in the linear regression model  using Huber's criterion has  been considered in \cite{owen2007robust,lambert2011robust,zoubir2018robust}. The derived MM  framework for  Huber's criterion can potentially lead to new research directions using theory developed in  \cite{schifano2010majorization,razaviyayn2013unified}.  
 
%-- Section 2 ---
\section{Maximum likelihood approach}  \label{sec:ML}

In his seminal work,  Huber \cite{huber1964robust} derived a family of univariate heavy-tailed distributions, which he called the ``least favorable distributions'' (LFDs).  
The LFD corresponds to a symmetric unimodal distribution which follows a Gaussian distribution in the middle, and a double exponential distribution in the tails. The standardized (unit scale) p.d.f. of LFD is  
 $f_0(x)=e^{- \rho_{c}(x) }$, 
 where  $\rho_{c}(x)$ is the \emph{Huber's loss function},  
\beq \label{eq:huber} 
\rho_{c}(x) =  \frac{1}{2} \times \begin{cases}  |x|^2, &\mbox{for  $|x| \leq c$} \\   2c |x| - c^2, &\mbox{for  $|x| > c$}, \end{cases} , \quad x \in \R,
\eeq 
where  $c$ is a  user-defined {\it threshold} that influences the degree of robustness.  Huber's loss \eqref{eq:huber}  is a hybrid of the (Least-squares) LS- and the least absolute deviation (LAD) loss functions using the LS-loss function for relatively small errors and LAD loss function for relatively large errors. 
Furthermore, in the limit $c \to\infty$, the loss function reduces to LS-loss $\rho_{c}(x)= \frac 1 2 x^2$. 
  Huber's loss function is  convex and differentiable, and the derivative of the loss function,
\[
\psi_{c}(x)=  \begin{cases} x, &\mbox{for  $|x| \leq c$} \\  c \,  \mathsf{sign}(x), &\mbox{for $|x|>c$}\end{cases}, 
\]
will be referred to as the \emph{score function} in the sequel. Note that $\psi_c(\cdot)$ is a winsorizing function. 

The threshold $c$ is usually chosen so that the minimizer in the regression-only problem ($\sigma$ being fixed) attains a user-defined 
 asymptotic relative efficiency (ARE)  w.r.t. the LS estimate (LSE) under Gaussian errors. 
 In order to obtain 95\% (or 85\%) ARE for the Gaussian noise case, the threshold are chosen according to $ c_{.95} =  1.345$  and $c_{.85}=0.7317$.

Assume now  that the error terms $e_i$ in the linear model \eqref{eq:linmodel} follow the LFD, so with the standardized distribution given previously. Then consider finding the ML estimates (MLE-s) of  the unknown parameters $\be$  and $\sig$ via  minimizing the negative log-likelihood
function of the data: 
\begin{equation} \label{eq:MLregscale}
  L_{\text{ML}} (\be,\sig)   = \ndim  \ln \sig + \sum_{i=1}^\ndim \rho_{c} \! \left(  \frac{y_{i} - \xr{i}^\top \be}{\sig} \right)  .
\end{equation}
 The problem is that the negative log-likelihood function is not convex in $(\be,\sig)$. This is easy to see by simply noting  that $L_{\textrm{ML}}(\be,\sig)$  is not convex in $\sig$ for a fixed $\be$.   Another problem is that the associated scale estimate will not be robust, e.g., possessing a bounded influence function.  
 The problem of  non-robustness of the associated MLE  and the non-convexity of the ML criterion \eqref{eq:MLregscale}, lead Huber to consider an alternative criterion function discussed in detail in the following sections.

%-- Section 3 ---
\section{Huber's criterion} \label{sec:huber} 

In order to avoid the problems associated  with \eqref{eq:MLregscale} described earlier, Huber \cite{huber1981robust} proposed an alternative criterion function 
\begin{align} \label{eq:Q}
 L (\be,\sig)  =    \ndim (\al \sig)  +  \sum_{i=1}^\ndim  \rho_c \! \left(  \frac{y_{i} - \xr{i}^\top \be}{\sig} \right) \sig  
\end{align} 
 where  $\al>0$ is a fixed scaling factor. We refer to  \eqref{eq:Q}  as {\it Huber's criterion} and the minimizer $(\hat \be,\hat \sig)$ of $L(\be,\sigma)$  as \emph{Huber's joint estimates of regression and scale}.   Unlike \eqref{eq:MLregscale}, Huber's criterion function \eqref{eq:Q} is  jointly convex in  $(\be,\sig)$.  
 In addition to this, the minimizer $\hat \be$ preserves the same theoretical robustness properties (such as bounded influence function) as  the estimator in the regression-only problem where the scale parameter is known. 

An important concept when optimizing the Huber's criterion is the pseudo-residual which is in essence a winsorized version of the conventional 
residual $\r= \y - \X \be$.  The  \emph{pseudo-residual} is  associated with the score function,  and defined as 
\beq  \label{eq:pseudo_residual} 
\r_\psi \equiv \r_\psi(\be,\sig)= \psi_c\! \left( \frac{\y-\X  \be}{\sigma}\right)  \sig, 
\eeq 
where $\psi_c(\r/\sig)= (\psi_c(r_1/\sig), \ldots, \psi_c(r_\ndim/\sig))^\top$.  
In the case of  LS-loss (the case $c \to  \infty$), $\psi_c(x)=x$, and $\r_\psi$ coincides with the conventional residual vector,  so $\r_\psi =\r$. 
Scaling by $\sig$ in  \eqref{eq:pseudo_residual}   maps the residuals back to the original scale of the data. 

Since the optimization problem \eqref{eq:Q} is convex, the global minimizer $(\hat \be,\hat \sig)$  is a stationary point of  \eqref{eq:Q}.  
Thus, $(\hat \be, \hat \sig)$ can be found by solving the \emph{$M$-estimating equations}, obtained by setting the gradient of $L(\be,\sig)$ to zero
\beq \label{eq:estim1} 
\begin{array}{lll} 
\nabla_{\be} L(\hat \be,\hat \sig)= 0  & \Leftrightarrow  & \x_j^\top   \hat \r_\psi    = 0 , \  \,  j=1,\ldots,p   \\
\nabla_{\sig} L(\hat \be,\hat \sig)= 0       &\Leftrightarrow  &  {\displaystyle \frac{1}{\ndim}\sum_{i=1}^\ndim  \chi_c\! \left( \frac{  y_i - \xr{i}^\top \hat \be  }{ \hat \sig} \right) = \al }
\end{array},
\eeq
 where   $\hat \r_\psi=\r_\psi(\hat \be,\hat \sig)$ and  $\chi_c: \R \to  \R_0^+$ is defined as
\begin{equation}
\chi_c(x) =    \psi_c(x)x - \rho_c(x) =    \frac{1}{2}  \psi_{c}(x)^2.   \label{eq:chi}
\end{equation}

It is instructive to consider the estimating equations \eqref{eq:estim1}  in the case of  LS loss $\rho_c(x)=\frac{1}{2}x^2$. 
In this case, the score function is $  \psi_c(x)=x$ and   $\hat \r_\psi= \hat \r$,   and the first equation in \eqref{eq:estim1} becomes  the conventional normal equations 
$
 \x_j^\top   \hat \r  = 0,  \ \mbox{ for } j=1,\ldots,p  \Leftrightarrow  \X^\top (\y -  \X \hat \be) = 0 .
$
 Hence,  the minimizer  $\hat \be$ of Huber's criterion is simply the LS estimate (LSE) of regression $\hat \be = (\X^\top \X)^{-1} \X^\top \y$ .  Furthermore,   the $\chi_c$ function in \eqref{eq:chi} is simply  $\chi_c(x)= \frac{1}{2} x^2$ and the 2nd estimating equation in \eqref{eq:estim1} reduces to $
\hat \sigma^2   = \frac{1}{\ndim (2 \al)} \sum_{i=1}^\ndim  ( \, y_i - \xr{i}^\top \hat \be)^2  . 
$
Thus, If we choose $\al = \frac{1}{2}$, then the solution $\hat \sigma$  coincides with the classical MLE of scale for Gaussian errors, the standard deviation of the residuals.    Interestingly,   the two quite different criterion functions, the Huber's criterion \eqref{eq:Q} and the ML criterion function \eqref{eq:MLregscale}, share the same  unique  global (joint) minimum when  when the LS-loss function is used. Therefore, Huber's criterion  can be seen as a method to convexify the Gaussian ML criterion function.    

The scaling factor $\al$  in  \eqref{eq:Q} is used to ensure  that   $\hat \sig$ 
is Fisher-consistent for the unknown scale $\sigma$ when $ \{ e_{i} \}_{i=1}^\ndim  \iidsim \mathcal N(0,\sigma^2)$.   Due to  \eqref{eq:estim1},   this is achieved if 
\begin{align}
\alpha  &=  \E[\chi_c(e)] = \frac{c^2}{2} (1-F_{\chi^2_1}(c^2)) + \frac 1 2 F_{\chi^2_3}(c^2),  \label{eq:alpha}
\end{align}
where $F_{\chi^2_k}$ denotes the c.d.f. of chi-squared distribution with $k$ degrees of freedom and  $ e \sim \mathcal N(0,1)$, 
Hence, when using the LS loss, $\chi_c(e)= \frac{1}{2} e^2$,  and  Fisher consistency is obtained if  $\al =   \frac 1 2 \E[ e^2] = \frac 1 2$

%-- Section 4 ---
\section{Minimization-Majorization Algorithm}
%--Subsection 4.0  (intro: could also be its own subsection)

A block-wise Minimization-Majorization algorithm works similarly to MM algorithms \cite{razaviyayn2013unified,hunter2004tutorial,sun2017majorization}. 
Let $\theta$ be partitioned into $\theta =(\theta_1,\theta_2)$, where $\theta_1  \in \Theta_1$,  $\theta_2  \in \Theta_2$ 
and $\Theta= \Theta_1 \times \Theta_2$ and we wish to find a minimizer of a real-valued function  $L(\theta) = L(\theta_1,\theta_2)$. 
 At $(n+1)$th iteration, the blocks are updated in a cyclic manner as follows: 
\begin{align}
\theta_2^{(n+1)}  &=  T_2 \big(\theta_1^{(n)},\theta_2^{(n)} \big), \label{eq:major2}  \\  &T_2 \big(\theta^{(n)}_1,\theta_2^{(n)} \big) 
= \underset{\theta_2 \in \Theta_2}{\mathrm{arg} \min} \,  g_2 \big(\theta_2 | \theta_1^{(n)}, \theta_2^{(n)} \big)  \notag \\
\theta_1^{(n+1)}  &=  T_1(\theta_1^{(n)},\theta_2^{(n+1)}),  \label{eq:major1}  \\   &T_1 \big(\theta^{(n)}_1,\theta_2^{(n+1)} \big) = \underset{\theta_1 \in \Theta_1}{\mathrm{arg} \min} \,  g_1 \big(\theta_1 | \theta_1^{(n)}, \theta_2^{(n+1)} \big)  \notag
 \end{align} 
 where the majorization surrogate functions $g_i(\theta_i | \theta_1',\theta_2')=g_i(\theta_i | \theta') $, $i \in \{1,2\}$ satisfy
\begin{align} 
g_i(\theta_i' | \theta_1',\theta_2')  &= L(\theta_1',\theta_2') \quad  \forall (\theta_1', \theta_2')  \in \Theta,  \label{eq:g1a}  \\ 
g_1(\theta_1 | \theta_1',\theta_2')  &\geq  L(\theta_1,\theta_2')   \quad  \forall \theta_1 \in \Theta_1, \forall (\theta_1', \theta_2')  \in \Theta, \label{eq:g2a} \\
g_2(\theta_2 | \theta_1',\theta_2')  &\geq  L(\theta_1',\theta_2)   \quad  \forall \theta_2 \in \Theta_2, \forall (\theta_1', \theta_2')  \in \Theta. \label{eq:g2b}
  \end{align} 
 Furthermore, when $ L$ and $g_i$ are differentiable functions, it is possible to impose the constraint 
\beq  \label{eq:cond_derivative}
\nabla_{\theta_i} g_i(\theta_i  \, | \,  \theta_1',\theta_2') \big|_{\theta_i= \theta_i'} = \nabla_{\theta_ i} L(\theta) \big|_{\theta=\theta'}, \ i \in \{ 1, 2\}.   
\eeq 
 Under some regularity conditions,  the sequence obtained by iterating the steps above  is a stationary point of $L(\theta)$ if it lies in
the interior of $\Theta$. 

  The challenging part in designing a block-wise MM algorithm is, naturally, in finding  
appropriate surrogate functions $g_i(\cdot | \theta')$, $i \in \{ 1,2\}$. 
A common choice is a quadratic function of the form $b_0 + b_1 \theta + b_2 \theta^2$ as it results in a simple
update formula in \eqref{eq:major1}, \eqref{eq:major2}.

We now construct an MM algorithm for obtaining the stationary solution  $(\hat \be,\hat \sig)$ in \eqref{eq:estim1} of Huber's criterion  $L(\be,\sig) = \sum_i \rho_c( r_i/\sig) \sig +   \ndim(\al \sig) $. For this purpose, we will let $\be'$ and $\sig'$ denote values of previous iterates, and write 
 $ \r' = \y - \X \be' $ and $\r'_\psi = \psi \big( \r'/\sig' \big) \sig'$ for the corresponding residual and pseudo-residual.  

For the scale term, we will use the following  majorization surrogate function:
 \beq \label{eq:maj_scale} 
 g_2(\sig | \be', \sig')  = a'  + b' \frac 1 \sig   +  \ndim \al \sig. 
 \eeq 
 In \eqref{eq:maj_scale} , $a'  + b'  \sig^{-1}$ is used to majorize $ \sum_i \rho_c( r_i/\sig) \sig $, where $a'$ and $b'$ are constants 
that depend on the previous iterates $\be'$ and $\sig'$. 
 These terms  can be found by solving the pair of equations  \eqref{eq:g1a} and \eqref{eq:cond_derivative} 
which yields (after simplifying)  the following surrogate function
\begin{align}
g_2(\sig |& \be',\sig') =L(\be',\sig')    \notag   \\ 
 &+ \ndim \al (\sig - \sig') + \sum_{i=1}^\ndim  \chi_c  \bigg(  \frac{  r_{i}' }{\sig'} \bigg) \bigg( \frac{ (\sig')^2}{\sig} - \sig' \bigg).  \label{eq:g2} 
 \end{align}
 Next we turn into  constructing a majorization surrogate function $g_1(\be | \be', \sig')$. For this purpose, consider a surrogate function for $\rho_c(r_{i}/\sig')$ by using 
 \[
 \rho_{\text{M}}\Big( \frac{r_i}{\sig'} \Big)  = a_{i}'  +   b_{i}'  \, \frac{r_i}{\sig'}    + \frac{1}{2}   \frac{r_i^2}{(\sig')^2} , 
\] 
where  the constants $a_{i}'$ and $b_{i}'$ depend on the previous iterates $\be'$ and $\sig'$ and are found by solving the pair of equations  \eqref{eq:g1a} and \eqref{eq:cond_derivative}  w.r.t. $a_{i}'$ and $b_{i}'$. After finding these solutions, we obtain a surrogate function of the form 
\begin{align*}
g_1(\be  | \be',\sig')  
&=  \ndim (\al \sig') +   \sum_{i=1}^\ndim \bigg ( a_{i}'   +   b_{i}'    \, \frac{r_{i}}{\sig'}  +  \frac{1}{ 2 } \frac{ r_{i}^2  }{(\sig')^2} \bigg) \sig'  \\
&=   \mbox{const} + \sum_{i=1}^\ndim \bigg(  \big[  r_{\psi,i}'  -  r_{i}'  \big]  r_{i}   + \frac{ 1 }{2} r_{i}^2   \bigg) \frac{1}{ \sig' }
 \end{align*}
where the constant term does not depend on $\be$.  
  
%--Subsection 4.1
\subsection{MM Algorithm for Huber's Criterion}  \label{sec:huber_mm_algor}

Next we prove that $g_1$ and $g_2$ are valid surrogate functions, so verify \eqref{eq:g2a} and  \eqref{eq:g2b}. 
First note that the difference function is 
\[
h_2(\sigma) = g_2(\sig | \be',\sig') - L(\be', \sig)  =  a_0 +   \frac{b_0 }{ \sigma}  - \sum_{i=1}^{\ndim} \rho_c\bigg( \frac{r'}{\sigma} \bigg)  \sigma
\]
for some constants $a_0$ and $b_0$.  Then note that   $h_2(\sigma)$ is a convex function in $1/\sigma$ since 
the first term is linear and  $ \rho_c(x)/x$ is a concave function in $x \geq 0$.  Furthermore, since $h_2(\sigma')=0$, it follows that $h_2(\sigma) \geq 0$, i.e., $g_2(\sig | \be',\sig') \geq  L(\be', \sig)$, for all $\sigma >0$.   This proves that $g_2$  verifies  \eqref{eq:g2b}. 
Next consider the difference function
\begin{align*}
&h_1(\be) = g_1(\be | \be',\sig')  -  L(\be,\sig')  \\
&=  \mbox{cnst} + \ \sum_{i=1}^\ndim \bigg(  \big[  r_{\psi,i}'  -   r_{i}'  \big]    r_{i}  + \frac{ 1 }{2} r_{i}^2 \bigg)\frac{1}{ \sig' } 
- \sum_{i=1}^\ndim \rho_c \bigg( \frac{r_i}{\sigma'} \bigg) \sigma'.
\end{align*} 
The Hessian matrix of the difference function $h_1$ is  
\begin{align*}
\bo H_1 &= \frac{\pr^2  h_1 }{\pr \be\pr \be^\top} 
 = \frac{1}{\sigma'} \sum_{i=1}^\ndim \bigg\{ 1 - \psi'_c \bigg( \frac{r_i}{\sigma'} \bigg) \bigg\} x_i x_i^\top  .
\end{align*}
Note that  $0 \leq \psi'_c(x) \leq 1$. Thus the matrix $\bo H_1$ is a positive semi-definite matrix, 
and thus the difference function $h_1$ is a convex function 
with a minimum at $\be'$. These results and the fact that $h_1(\be')= 0$ imply that $g_1(\be | \be',\sig')  \geq  L(\be,\sig') $,   $\forall \be$.  
In the next theorem  we obtain the minimizers.  The proof is omitted due to lack of space.

\begin{theorem}  \label{th:MMupdates} 
The MM update of scale is 
\begin{align} 
\sig^{(n+1)}
 &=  \underset{\sig>0}{\mathrm{arg} \, \mathrm{min}}  \, g_2(\sig | \be^{(n)},\sig^{(n)})  = \sig^{(n)} \tau  ,
 \end{align} 
 where  
 \beq \label{eq:tau} 
 \tau =   \frac{1}{ \sqrt{2 \ndim \alpha} } \bigg \| \psi_c  \bigg(   \frac{\r^{(n)}}{\sig^{(n)} }\bigg)  \bigg\| 
 \eeq 
and $ \r^{(n)} =  \y - \X \be^{(n)}$.
The MM update for regression is 
\begin{align*} 
\be^{(n+1)} &= \underset{\be \in \R^{p+1}}{\mathrm{arg} \, \mathrm{min}}  \, g_1(\be | \be^{(n)},\sig^{(n+1)})  
=  \be^{(n)} +   \delta  ,
\end{align*} 
where 
\beq  \label{eq:delta} 
 \delta =   \X^+ \psi_c \bigg( \frac{\r^{(n)}}{\sig^{(n+1)} } \bigg) \sig^{(n+1)}
 \eeq 
and $\X^+= (\X^\top \X)^{-1} \X^\top$.
 \end{theorem}

 The  updates of \autoref{th:MMupdates}  thus form the basic frame for the MM algorithm that is described in \autoref{alg:HubReg}.

\begin{algorithm}[!t]
\SetVlineSkip{0pt}      
\IncMargin{0.3em}
\DecMargin{2em}
\SetNlSkip{0.7em}
\SetKwInOut{Input}{input}
\SetKwInOut{Output}{output}
\SetKwInOut{Init}{initialize}
\Input{ $(\y,\X)$, threshold $c>0$,  initial guess $(\be^{(0)},\sig^{(0)})$,  
 $\epsilon>0$, $\mu^{(0)} = 0$ and $\lam^{(0)}=1$}
\Init{$N_{\mathrm{iter}} \in \mathbb{N}$, compute $\al=\al(c)$ in \eqref{eq:alpha} and  $\X^+$} 
 
 \BlankLine 

\nonl \For{$n=0,1,\ldots, N_{\mathrm{iter}}$ }{
 \label{InRes1} 

Update the residual
$ \r^{(n)} = \y-\X \be^{(n)}$  and $\tau$  in \eqref{eq:tau}

Update the step size for scale: \label{ln:step_size_scale} 
\begin{align*} 
\lam^{(n+1)} =  \lam^{(n)} + \frac{ \log  \left\| \psi_c \Big(  \dfrac{\r^{(n)}}{\sig^{(n)} \tau^{\lam^{(n)}}} \Big)  \dfrac{1}{ \sqrt{2 \al \ndim}}\right\| }{ \log \tau}
\end{align*}

Update  scale $\sig^{(n+1)} = \sig^{(n)} \tau^{\lambda^{(n+1)}}$ and  $\delta$ in \eqref{eq:delta}
  
Update the step size for regression: \label{ln:step_size_reg} 
\begin{align*} 
\bo z &= \X  \delta,\quad \bo w = w \bigg( \frac{ \r^{(n)} - \mu^{(n)} \bo z}{\sig^{(n+1)}} \bigg) \\
 \mu^{(n+1)} &=   \| \bo z  \|_{\bo w }^{-2}  \langle \r^{(n)} ,  \bo z \rangle_{\bo w } 
 \end{align*}

Update the regression vector:
\begin{center} $ \be^{(n+1)} =  \be^{(n)} +  \mu^{(n+1)}  \delta $ \end{center}

\If {$\|  \mu^{(n+1)}  \delta \|/\|  \be^{(n)} \| < \epsilon$  \mbox{ \&} $|\tau^{\lambda^{(n+1)}}-1|< \epsilon$}{
\nonl  \Return  $(\hat \be,\hat \sig) \leftarrow (\be^{(n+1)},\sig^{(n+1)})$ }
 }
 \Output{$(\hat \be, \hat \sig)$,  the minimizer of  $ L(\be,\sig)$.}

 \caption{ \texttt{hubreg}, solves \eqref{eq:Q}  via  MM.}
\label{alg:HubReg}
\end{algorithm}
 
%--Subsection 4.2
\subsection{Step size selection}

An additional change to the MM algorithm is  the introduction of the step sizes in \autoref{ln:step_size_scale}  and 
\autoref{ln:step_size_reg}. 

 Since the update for $\be$ can be viewed as a gradient descent 
move towards the direction  $\delta$,  one may try to identify an optimal step size that maximally
reduces the objective function at each iteration. In other words, we minimize Huber's criterion  $L(\be,\sig)$ with  $\sig=\sig^{(n+1)}$ and  
 $\be$ fixed at  $\be=\be^{(n)} + \mu  \delta$: 
\begin{align}  
\mu^{(n+1)} &= \argmin_{\mu}  \sum_{i=1}^\ndim \rho_c  \bigg( \frac{y_i - x_i^\top ( \be^{(n)} + \mu  \delta)}{ \sigma^{(n+1)}}  \bigg)  \notag \\ 
&= \argmin_{\mu}  \sum_{i=1}^\ndim \rho_c  \bigg( \frac{ r_i^{(n)}  - \mu  x_i^\top \delta}{ \sigma^{(n+1)}} \bigg) \label{eq:reg_simple} .
\end{align} 
Solving \eqref{eq:reg_simple}  is equivalent to computing Huber's $M$-estimator of regression with auxiliary scale $\sigma^{(n+1)}$ in the simple (one predictor) linear  regression model with response vector $\r^{(n)}$ and  regressor $\bo z= \X  \delta$. A standard approach  for finding the (unique) minimizer of the convex optimization problem in  \eqref{eq:reg_simple}  is the \emph{iteratively reweighted LS (IRWLS) algorithm} \cite{maronna2006robust}, which iterates the steps
\beq \label{eq:IRWLS}
\begin{aligned} 
 \bo w &\leftarrow w \left( \frac{ \r^{(n)}  - \mu \bo z }{\sig^{(n+1)}} \right) , \quad 
  \mu &\leftarrow   \| \bo z  \|_{\bo w }^{-2}  \langle \r^{(n)} ,  \bo z  \rangle_{\bo w} 
\end{aligned}
\eeq
until convergence, given an initial value of $\mu$ to start the iterations. Here $ \langle \bo a ,  \bo b  \rangle_{\bo w}  =  \sum_{i=1}^\ndim a_ib_i w_i$ and 
$ \| \bo a \|_{\bo w}^2 =  \langle \bo a ,  \bo a  \rangle_{\bo w}$. 
Instead of iterating until convergence, we use a 1-step estimator, which correspond to a single iteration of  \eqref{eq:IRWLS} with initial value given by previous value of step size $\mu^{(n)}$. 
This results to the update shown in \autoref{ln:step_size_reg} of  \autoref{alg:HubReg}.

Next we turn our attention to the step size for scale. First we note that the update $\sigma^{(n+1)} = \sigma^{(n)} \tau$ can be 
viewed  as a gradient descent move in the log-space (after change of variables $a=\log \sigma$, i.e, $\sigma=e^a$), 
$a^{(n+1)} = a^{(n)}   +  \lam \log (\tau )$ 
with stepsize $\lam=1$.  Thus employing a data adaptive step size $\lambda = \lambda^{(n+1)}$ translates to update 
$\sigma^{(n+1)} = \sigma^{(n)} \tau^{\lam^{(n+1)}}$. 
We then identify an optimal step size that maximally
reduces the objective function in each iteration. In other words, we minimize Huber's criterion  $L(\be,\sig)$ with 
 $\be$ fixed at  $\be=\be^{(n)}$ and  $\sig=\sig^{(n)} \tau^\lam$:  
\begin{align}  \label{eq:scale_simple} 
\lam^{(n+1)} = \argmin_{\lam} \, \ndim \al  \tau^\lam +   \sum_{i=1}^\ndim \rho_c  \bigg( \frac{\tilde r_i^{(n)}}{ \tau^\lam}  \bigg)  \tau^\lam  
\end{align} 
where $\tilde r_i^{(n)}=  r_i^{(n)}/\sig^{(n)} = (y_i - x_i^\top  \be^{(n)})/\sig^{(n)}$.  As was shown in \autoref{th:MMupdates}, an MM algorithm that  finds the optimum of \eqref{eq:scale_simple}  would iterate  
\begin{align*}  
\tau^{\lam} \gets  \tau^\lam  \left\|  \psi_c \bigg( \frac{\tilde \r^{(n)} }{ \tau^{\lam}} \bigg)  \frac{1}{ \sqrt{2 N \al}}  \right\|
\end{align*} 
starting from some initial value for $\lam$ until convergence. The iteration update above can be written equivalently as 
\begin{align} \label{eq:lam_update}   
\lam \gets  \lam  +  \log  \left\|  \psi_c \big( \tilde \r^{(n)}/\tau^{\lam} \big)  \frac{1}{ \sqrt{2 N \al}}  \right\|/ \log \tau.
\end{align} 
Again, instead of iterating \eqref{eq:lam_update}    until convergence, we use a 1-step estimator, which correspond to a single iteration of  \eqref{eq:lam_update}   with initial value given by previous value of step size $\lam^{(n)}$. 
This results to the update shown in \autoref{ln:step_size_scale} of  \autoref{alg:HubReg}.

%-- Section 5 --
\section{Sparse learning via Huber's criterion}  \label{sec:huber_sparse} 

Next consider a sparse signal recovery (SSR) problem \cite{donoho2006stable,duarte2011structured}, 
where $\be$ in  \eqref{eq:linmodel}  is $K$-sparse, i.e., the \emph{support} 
 $ \Gamma= %\mathsf{supp}(\be)= 
 \{   i \in \{1,\ldots, p \} \, : \: \beta_{i} \neq 0  \} $ 
 has at most $K$-nonzero elements, so $\| \be \|_0 = | \Gamma | \leq K$.  Furthermore, the dimensionality $p$ can be greater than the number of measurements $N$. This means that the design matrix $\X$ can be  underdetermined (and not of full rank).  In the context of SSR, $\X$ is often referred to as \emph{measurement matrix} or \emph{dictionary} and $\be$ as the \emph{signal vector}.  
 
 Traditional non-robust approaches aim at minimizing $ \| \y - \X \be \|^2$ w.r.t. $\be$ subject to $\| \be \|_0 \leq K$. This   is a combinatorial NP-hard problem. However,  when the measurement matrix $\X$ satisfies  certain coherence conditions, bounds on
the recovery error are known for several reconstruction algorithms when the measurements are corrupted by noise with bounded norm. 
One such algorithm is the \emph{normalized iterative hard-thresholding (NIHT)} algorithm \cite{blumensath2010normalized}. 

The greedy NIHT approach  for minimizing $L(\be,\sig)$ subject to  $\| \be \|_0 \leq K$ iterates the following steps \cite{ollila2014robust}: 
\begin{enumerate} 
\item update $\r^{(n)} = \y-\X \be^{(n)}$ and  $\tau$  in \eqref{eq:tau}. 
\item update the step size $\lam^{(n+1)}$ for scale. 
%Update the scale
\item update the scale $\sig^{(n+1)} = \sig^{(n)} \tau^{\lambda^{(n+1)}}$.
%Re-update the pseudo-residual 
\item update the step size  $\mu^{(n+1)}$ for signal vector.
\item update  $\r_\psi^{(n+1)} = \psi_{c}  \bigg( \dfrac{\r^{(n)}}{\sig^{(n+1)}} \bigg) \sig^{n+1}$.
\item $\be^{(n+1)} =  H_K \big( \be^{(n)} +  \mu^{(n+1)}  \X^\top r_\psi^{(n+1)} \big)$.
\end{enumerate} 
until convergence, where $H_K(\be)$ denotes the \emph{hard thresholding operator}: it retains the $K$ elements of  vector $\be$ that are largest in absolute value and set the other elements to zero.  The step size for scale and regression can be found as described in \autoref{alg:HubReg}.  This robust SSR method described above is the same as in \cite{ollila2014robust}  except we have incorporated the proposed step size for scale. We refer to this method  as HUBNIHT  in the sequel.  Extension of  \cite{ollila2014robust} to 
\emph{simultaneous} SSR  \cite{duarte2011structured}  problem has been covered in  \cite{ollila2015robust} and \cite{ollila2015multichannel} for real- and complex-valued signals, respectively (and referred to as HUBSNIHT).

%-- Section 6 --
\section{Applications} 
\subsection{Regression example} 

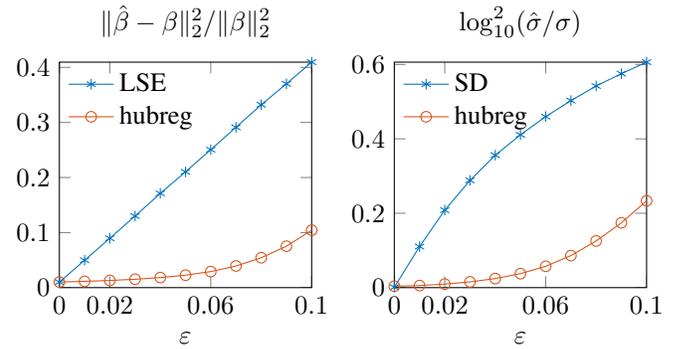
\begin{figure}[h]
	\centering
\definecolor{mycolor1}{rgb}{0.00000,0.44700,0.74100}%
\definecolor{mycolor2}{rgb}{0.85000,0.32500,0.09800}%
\begin{tikzpicture}

\begin{axis}[%
width=0.39\columnwidth,
height=3cm,
at={(0.969in,0.747in)},
scale only axis,
xtick={0, 0.02, 0.06, 0.1},
xmin=0,
xmax=0.1,
xlabel={$\varepsilon$},
ymin=0,
ymax=0.409495100444616,
title={$\| \hat{\beta} - \beta \|^2_2/\| \beta \|^2_2$},
xticklabel style={/pgf/number format/fixed},
legend style={at={(0.001,1)}, anchor=north west, legend cell align=left, align=left, draw=none, fill=none}
]
\addplot [color=mycolor1, mark=asterisk, mark options={solid, mycolor1}]
  table[row sep=crcr]{%
0	0.0100374628631225\\
0.01	0.0496761352894945\\
0.02	0.0896305353794545\\
0.03	0.13001143683751\\
0.04	0.171131600940637\\
0.05	0.210256235165709\\
0.06	0.250515559672096\\
0.07	0.291173897977728\\
0.08	0.332117524120715\\
0.09	0.370132330833543\\
0.1	0.409495100444616\\
};
\addlegendentry{LSE}

\addplot [color=mycolor2, mark=o, mark options={solid, mycolor2}]
  table[row sep=crcr]{%
0	0.0101605709558678\\
0.01	0.0114059666627534\\
0.02	0.0130425440475739\\
0.03	0.0151965189901644\\
0.04	0.018273065458072\\
0.05	0.0225926670309674\\
0.06	0.0291243225752347\\
0.07	0.0392895227301099\\
0.08	0.0542675611477762\\
0.09	0.0752077108508448\\
0.1	0.104363110947723\\
};
\addlegendentry{hubreg}

\end{axis}

\end{tikzpicture}%
~%
\begin{tikzpicture}

\begin{axis}[%
width=0.39\columnwidth,
height=3cm,
scale only axis,
xmin=0,
xmax=0.1,
xtick={0, 0.02, 0.06, 0.1},
xlabel={$\varepsilon$},
ymin=0,
ymax=0.606283915318875,
xticklabel style={/pgf/number format/fixed},
title={$\log_{10}^2(\hat{\sigma}/\sigma)$},
legend style={at={(0.001,1)}, anchor=north west, legend cell align=left, align=left, draw=none, fill=none}
]
\addplot [color=mycolor1, mark=asterisk, mark options={solid, mycolor1}]
  table[row sep=crcr]{%
0	0.00109908319766168\\
0.01	0.110338293044071\\
0.02	0.20836448264589\\
0.03	0.288370840484129\\
0.04	0.355983340558221\\
0.05	0.410846191125821\\
0.06	0.459824984863869\\
0.07	0.50288893605216\\
0.08	0.542170149336636\\
0.09	0.575276250538618\\
0.1	0.606283915318875\\
};
\addlegendentry{SD}

\addplot [color=mycolor2,mark=o, mark options={solid, mycolor2}]
  table[row sep=crcr]{%
0	0.00386247645539199\\
0.01	0.00557728677115272\\
0.02	0.00951207499175939\\
0.03	0.015416763061439\\
0.04	0.0243747956031675\\
0.05	0.03753027222736\\
0.06	0.0573324211537381\\
0.07	0.086177349918668\\
0.08	0.125453879860899\\
0.09	0.174371177343465\\
0.1	0.233569099998435\\
};
\addlegendentry{hubreg}

\end{axis}
\end{tikzpicture}%
	\vspace{-2.5em}
	\caption{Evolution of estimation error versus sign-change error probability $\varepsilon$ of  outputs $y_i$; $N=500$, $p=250$, $\mathrm{SNR}=20\mathrm{dB}$, $c=1.345$, $2000$ trials.}
	\label{fig: regression_examples}
\end{figure}

First we consider a simple regression experiment to validate the proposed MM framework for Huber's criterion. We consider a linear model as in \eqref{eq:linmodel} where $N=500$, $p=250$, the $e_i$ are generated through standardized Gaussian distribution and $\sigma$ is chosen so that the signal to noise ratio is $\mathrm{SNR}=20\mathrm{dB}$. Additionally, some errors are introduced into the model by changing the sign of random samples $y_i$ chosen according to a Bernoulli distribution with probability $\varepsilon \in [0,1]$. We then compare the performance of the hubreg estimator ($c=1.345$)  of \autoref{alg:HubReg}   compared to the LSE for regression and standard deviation (SD) for scale by performing $2000$ Monte-Carlo trials. The results displayed in \autoref{fig: regression_examples} show that while the estimation is correct for both methods when there is no outliers, the hubreg estimator is more robust to the presence of errors.

\vspace{-1em}
\vspace{-1em}

%--Subsection 6.2 (image denoising) 
\subsection{Image denoising} 

To illustrate potential applications of the sparse framework presented in \autoref{sec:huber_sparse}, we consider an image denoising problem of a grayscale image. Following a sliding windows approach, we consider the pixel values $\mathbf{y}$ on the local patch as the outputs of a linear model where the inputs $\mathbf{X}$ constitutes an overcomplete dictionary. Such dictionaries contain more atoms than their dimensions and are often redundant so a valid representation of the image must be sparse. Given this dictionary, an encoding $\hat{\beta}$ can be obtained by minimizing Huber's criterion with $K$-sparsity constraint using the HUBNIHT algorithm. Thanks to the robustness of this criterion to outliers, as shown previously, it is possible to reconstruct pixels values $\hat{\mathbf{y}}=\mathbf{X}\hat{\beta}$ that are less affected by the noise.

\begin{figure}[h]
	\centering
	\includegraphics[width=0.24\columnwidth, trim={2cm 1.5cm 2cm 1cm}, clip]{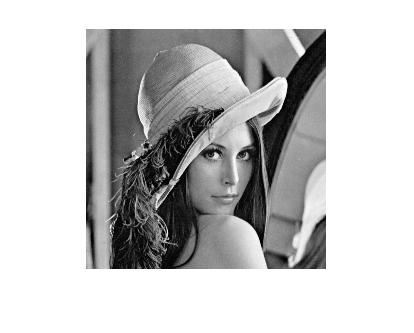}
	\includegraphics[width=0.24\columnwidth, trim={2cm 1.5cm 2cm 1cm}, clip]{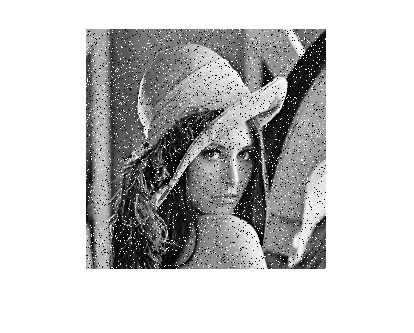}
	\includegraphics[width=0.24\columnwidth, trim={2cm 1.5cm 2cm 1cm}, clip]{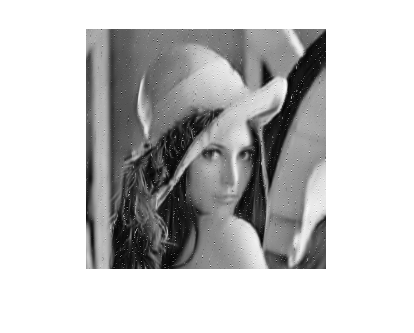}
	\includegraphics[width=0.24\columnwidth, trim={2cm 1.5cm 2cm 1cm}, clip]{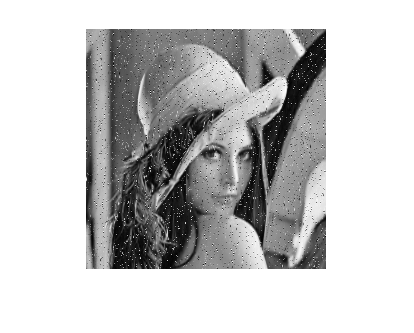}\\
	\hfill i)\hfill ii)\hfill iii)\hfill iv) \hfill
	\caption{i) Original Image. ii) Noisy image. iii) Reconstructed image ($K=6$). iv) Reconstructed image ($K=10$).}
	\label{fig: image denoising}
\end{figure}

An example of denoising via sparse reconstruction by using Huber's criterion is given in \autoref{fig: image denoising}. The dictionary, of size $p=192$, has been constructed using standard Debaucheries and Coiflets dictionaries often used in image processing applications 
and a window of size $8\times 8$ has been used. The difference between the two obtained results highlight that there is a compromise between sparsity to reduce noise and loss of details in the original image.

%--Subsection 6.3 (dictionary learning) 
\subsection{Dictionary learning} 

Finally, let us consider the problem of dictionary learning which aims at finding a sparse representation as in the SSR problem but also aims to learn the dictionary directly from the data. This can be formulated as the following optimization problem:
\begin{equation}
\min_{\be, \mathbf{X}\in\mathcal{D}} \| \y - \X \be \|^2 \quad \mbox{subject to} \quad \| \be \|_0 \leq K,
\label{eq: dictionary learning}
\end{equation}
where $\mathcal{D}$ is a set imposing constraints to the dictionary elements in order help resolve the scale invariance problem between $(\beta, \mathbf{X})$. Among those constraints, unit $\ell_2$-norm of each column of $\mathbf{X}$ is often used in the literature.	
Traditionally algorithms that solve \eqref{eq: dictionary learning} iterate between a step of sparse coding, where $\mathbf{X}$ being fixed, and a step of estimation of dictionary $\mathbf{X}$, where $\beta$ is fixed (obtained in the sparse coding step). Popular algorithms include the Method of Optimal Directions (MOD) \cite{engan99method} or the K-SVD \cite{aharon2006ksvd} algorithms. Both of the methods perform sparse coding using the matching pursuit algorithm and differ in the way the dictionary is learned from the data. Since the  MM-framework presented in this paper has been successfully applied to sparse coding, it can replace the matching pursuit used  in both approaches. This plug-in methodology is expected to inherit robustness properties from Huber's estimates. This research direction will be studied in the subsequent extended journal version of this work.  

%\vfill
%\pagebreak


\end{document}